\documentclass[lettersize,journal]{IEEEtran}

\usepackage[utf8]{inputenc} % allow utf-8 input
\usepackage[T1]{fontenc}    % use 8-bit T1 fonts

\usepackage{moreverb,url}
\usepackage{amsmath,amssymb,amsfonts}
\usepackage{algorithmic}
\usepackage{array}
\usepackage{mathtools}
\usepackage[caption=false,font=normalsize,labelfont=sf,textfont=sf]{subfig}
\usepackage{textcomp}
\usepackage{stfloats}
\usepackage{url}
\usepackage{pifont}
\usepackage{graphicx}
\usepackage{balance}

\usepackage[table]{xcolor}
\definecolor{red}{rgb}{1.00,0.00,0.00}
\definecolor{blue}{rgb}{0.00,0.00,1.00}
\definecolor{green}{rgb}{0.30, 0.50,0.00}
\definecolor{lightgray}{gray}{0.9}
\definecolor{purple}{rgb}{0.60,0.10,0.90}

\usepackage{cite}

\usepackage{mathtools}
\usepackage{proof}

\usepackage{tabularx}
\usepackage{booktabs}
\usepackage{multicol}
\usepackage{multirow}
\usepackage{floatrow}
\usepackage{float}
\usepackage{url}
\usepackage{hyperref}
\usepackage{tikz}
\usepackage{ctable}
\usepackage{pgf}
\usetikzlibrary{shapes,external}
\usetikzlibrary{shapes.multipart}
\usetikzlibrary{calc, positioning, automata}

\definecolor{codepurple}{rgb}{0.58,0,0.82}

\title{\LARGE \bf
Parse-Augment-Distill: Learning Generalizable Bimanual \\ Visuomotor Policies from Single Human Video
}

\author{Georgios Tziafas$^{1*}$, Jiayun Zhang$^{1}$, Hamidreza Kasaei$^{1}$% <-this % stops a space
\thanks{$^{*}$Corresponding Author  \{ {\tt\small g.t.tziafas@rug.nl} \} \newline
        $^{1}$ Department of Artificial Intelligence, University of Groningen, the Netherlands
}}

\begin{document}

\maketitle
\thispagestyle{empty}
\pagestyle{empty}

%%%%%%%%%%%%%%%%%%%%%%%%%%%%%%%%%%%%%%%%%%%%%%%%%%%%%%%%%%%%%%%%%%%%%%%%%%%%%%%%
\begin{figure*}[!b]
  \centering
  \includegraphics[width=\textwidth]{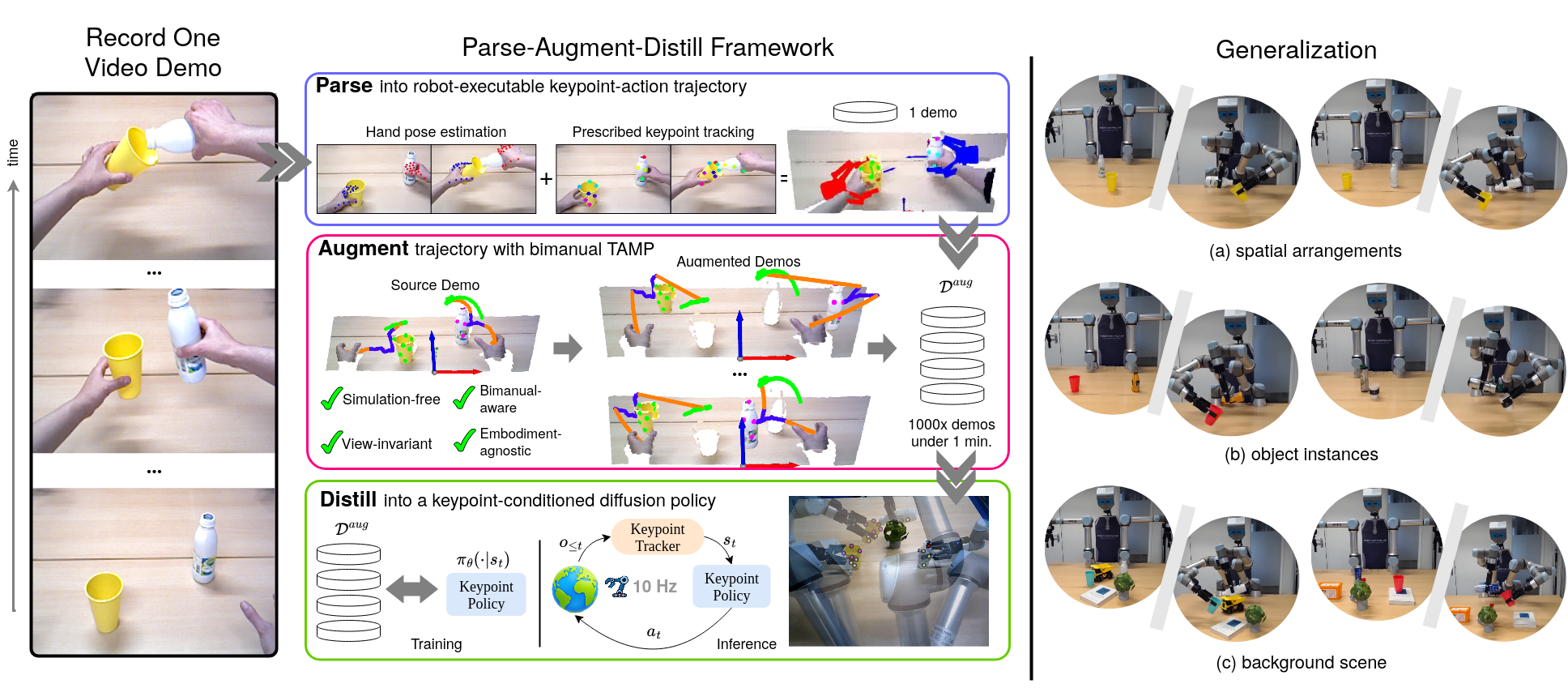}
  \caption{\textbf{PAD framework overview:} Given one human video demonstration, PAD executes three subsequent steps: (a) parsing the video into a robot-executable state-action trajectory, (b) spatially augmenting the demo at scale via bimanual TAMP, and (c) distilling the generated data into a closed-loop keypoint policy. The obtained policy can generalize to unseen spatial arrangements, object instances and background scene noise.}
  \label{fig:fig1}
\end{figure*}

\begin{abstract}
Learning visuomotor policies from expert demonstrations is an important frontier in modern robotics research, however, most popular methods require copious efforts for collecting teleoperation data and struggle to generalize out-of-distribution. Scaling data collection has been explored through leveraging human videos, as well as demonstration augmentation techniques. The latter approach typically requires expensive simulation rollouts and trains policies with synthetic image data, therefore introducing a sim-to-real gap. In parallel, alternative state representations such as keypoints have shown great promise for category-level generalization. In this work, we bring these avenues together in a unified framework: PAD (Parse-Augment-Distill), for learning generalizable bimanual policies from a single human video. Our method relies on three steps: (a) parsing a human video demo into a robot-executable keypoint-action trajectory, (b) employing bimanual task-and-motion-planning to augment the demonstration at scale without simulators, and (c) distilling the augmented trajectories into a keypoint-conditioned policy. Empirically, we showcase that PAD outperforms state-of-the-art bimanual demonstration augmentation works relying on image policies with simulation rollouts, both in terms of success rate and sample/cost efficiency.
We deploy our framework in six diverse real-world bimanual tasks such as pouring drinks, cleaning trash and opening containers, producing one-shot policies that generalize in unseen spatial arrangements, object instances and background distractors.
Supplementary material can be found in the project webpage \href{https://gtziafas.github.io/PAD_project/}{\textcolor{codepurple}{https://gtziafas.github.io/PAD\_project/}}.

\end{abstract}

%%%%%%%%%%%%%%%%%%%%%%%%%%%%%%%%%%%%%%%%%%%%%%%%%%%%%%%%%%%%%%%%%%%%%%%%%%%%%%%%
\section{INTRODUCTION}
Visuomotor policy learning for robot manipulation has seen great success in recent years \cite{DiffusionP, MobileAL, zhao2023learningfinegrainedbimanualmanipulation, RDT, Kim2024OpenVLAAO}, yet it typically demands costly and time-consuming data collection from expert demonstrators.
This data-hungriness stems from the different required axes of generalization: a competent policy must generalize in unseen object arrangements (\textit{spatial}) and object instances (\textit{object}), as well as be robust to environmental conditions such as scene background, camera placement etc. (\textit{background}).
As a result, most common policies struggle to generalize in out-of-distribution scenarios where corresponding data has not been collected.
A recent methodology to tackle this data scarcity is to tap into the vast repository of videos available in the web, showcasing humans interacting with objects in diverse scenarios \cite{Ego4D,videodataset1,videodataset2,videodataset3,videodataset4}.
Here the main challenge is bridging the embodiment gap between humans and robot morphologies \cite{Ren2025MotionTA,Phantom,PointPolicy}.
Alternatively, a recent line of works aims at dealing with spatial generalization by augmenting a small number of source demos with structured, object-centric \textit{task-and-motion planning (TAMP)} procedures \cite{MimicGen,IntervenGen,SkillMimicGen,DexMimicGen}.
However, most works train image policies that require calibrated digital twins and expensive on-robot rollouts to generate the augmentations, therefore introducing a visual sim-to-real gap, while still struggling with object and background generalization.
When it comes to bimanual manipulation, additional considerations related to arm collaboration strategies for different task scenarios further complicate data collection / generation.

% \paragraph{General} General story about how inefficient end-to-end behavioral cloning is. Mention ways to alleviate this: (a) Using human video data, (b) Demo augmentations, (c) Keypoint state representations.

% In this work we wish to unify these avenues into a single framework, termed \textbf{PAD} (\textbf{P}arse \textbf{A}ugment \textbf{D}istill).
% Explain what the framework does in high-level.
 In this work we wish to tackle these challenges by proposing \textbf{PAD} (\textbf{P}arse-\textbf{A}ugment-\textbf{D}istill), a unified framework for learning bimanual visuomotor policies from a single human video demonstration.
Our framework works in three steps (see Fig.~\ref{fig:fig1}): (a) parsing the video into robot-executable data, (b) augmenting the data in a simulation-free fashion and, (c) distilling the augmented data into a closed-loop policy.

In our work, we explicitly seek spatial, object and background generalization.
To accommodate this, we utilize 3D keypoint coordinates as state representations for our trained policy, which offers three important advantages:
First, keypoints abstract the visual scene into a low-dimensional geometric representation, which is task-specific and decoupled from object semantics, and therefore has empirically shown to aid in sample-efficiency and robustness to background noise \cite{KALM,PointPolicy,P3PO}.
Second, keypoints facilitate category-level object generalization, inherited by the open-world capabilities of pretrained vision models for identifying semantic correspondences \cite{DINO, DIFT, SDDINOv2}.
Finally, 3D point states enable efficient spatial augmentations, as keypoint coordinates can be computed on-the-fly through 3D rigid geometry assumptions \cite{DemoGen}.
This alleviates the need for a digital twin and expensive simulation rollouts, which would be required by a typical image policy to obtain image observations \cite{MimicGen,IntervenGen,SkillMimicGen,DexMimicGen}.
In turn, this significantly improves data collection time and bridges the sim-to-real gap introduced by simulators.

Concretely, in PAD we introduce a general TAMP framework for spatial demo augmentations, specialized for bimanual manipulation.
To that end, we introduce bimanual task templates, symbolic representations that declare information about each arm's object assignments, involved contacts and requirements for arm synchronization, while abstracting away the specific semantics of the task.
We particularly focus on handling issues related to bimanual manipulation, such as out-of-range arm-object assignments and re-synchronization between the arms during motion planning, which are missing from previous works in bimanual demo augmentation \cite{DexMimicGen}.
Our augmentation framework is general, cost-efficient and embodiment-agnostic, as it uses human video as the source demo that can be mapped to any given morphology.
Finally, we use prescribed 3D keypoints as our state representation instead of RGB images or point-clouds, and accompany them with augmentations that aid the policy in object generalization.
To distill the augmented data, we introduce Kp-RDT, an adapted version of RDT \cite{RDT} for learning bimanual diffusion policies with keypoint conditioning.
% \paragraph{Emphasize contribution of dataset augmentation} Extend previous works to bimanual TAMP. Arm synhcronization. Keypoint augmentations.

% \paragraph{Summarize contributions bullet list} 
Empirically, we show that our framework outperforms state-of-the-art bimanual demo augmentation methods \cite{DexMimicGen} in four simulation tasks from the DexMimicGen \textit{robosuite} benchmark \cite{robosuite}, both in terms of success rates, as well as sample-efficiency and data generation time.
We further apply our framework in six diverse real-world tasks and show that PAD obtains policies that generalize to unseen spatial arrangements, object instances and background scene noise, while doing so from a single human demonstration.

In summary, our contributions with this work are threefold:
\begin{itemize}
    \item We introduce PAD, a unified framework for generalizable bimanual policy learning from a human video. 
    \item We propose a general bimanual TAMP framework for spatial demo augmentations, applicable to a wide variety of manipulation skills and arm-coordination strategies, as well as open-ended object categories.
    \item We perform extensive robot experiments in 10 tasks, 4 in simulation and 6 with hardware, demonstrating significant gains compared to previous works in terms of success rates and sample/cost efficiency, as well as strong generalization in real-world tasks.
\end{itemize}

\section{RELATED WORK}

\noindent \textbf{Learning from egocentric video.} 
% There have been several attempts at learning robot policies from human videos.
Many recent datasets \cite{Ego4D,videodataset1,videodataset2,videodataset3,videodataset4}
represent large-scale efforts to collect egocentric hand-object interaction videos in diverse real-world scenes.
Such datasets have been used for learning robotics-tailored visual representations as a pretraining step \cite{Bhateja2024RoboticOR,r3m,Wu2023UnleashingLV,Ma2022VIPTU,Ma2023LIVLR,Karamcheti2023LanguageDrivenRL}.
Other works learn coarse policies from human videos, using keypoints \cite{Bharadhwaj2024Track2ActPP} or generative modeling \cite{Bharadhwaj2024Gen2ActHV}, which will later be fine-tuned for specific embodiments and tasks.
The recent work MT-$\pi$ \cite{Ren2025MotionTA} proposes to co-train a keypoint policy with human videos and robot data simultaneously.
All the above lines of work still require robot data to obtain a performant policy, typically collected through costly teleoperation.
Phantom \cite{Phantom} proposes to edit human videos with generative models to map them into robot data, while other works propose to utilize hardware platforms such as smart glasses \cite{EgoMimic} or MoCaP-like setups \cite{DexMV} to collect video data for dexterous manipulation.
Recent works \cite{P3PO,PointPolicy,liu2025egozerorobotlearningsmart} unify human video and robot data through point-based representations for both objects and hands.
% All above works is that despite having access to abundant human demonstrations, there is a need to collect robot data to achieve a highly performant policy. 
However, these works do not consider demo augmentations and hence require a lot of videos to train policies that generalize.
In our work we introduce demo augmentations for bimanual manipulation and obtain generalizable policies from a single human video.

\noindent \textbf{Keypoint-based representations.} Finding correspondences between objects \cite{Lai2021TheFC} has been extensively explored for robotics applications, either analytically \cite{Rodriguez2018TransferringCF,Biza2023OneshotIL} or with data-driven methods \cite{kPAM,kPAM2,Turpin2021GIFTGI,Wen2022YouOD,Simeonov2021NeuralDF}.
Although data-driven approaches have shown better generalization, they require additional training data such as 3D object shape or labeled keypoint datasets.
More recently, several works have shown that the emergent capabilities of modern vision models \cite{DIFT, SDDINOv2, DINO} for identifying keypoint correspondences in-the-wild serves as a powerful proxy for category-level object generalization \cite{DINOBot,Robo-ABC,SparseDFF,D3Fields,Goodwin2023YouOL,Hadjivelichkov2022OneShotTO}. 
The recent work KALM \cite{KALM} utilizes multimodal large language models (MLLMs) \cite{Gemini, GPT4} to automatically discover keypoint annotations in robot data for training and employs open-loop keypoint-conditioned policies that show good generalization.
All above works need to collect robot demonstrations for specific tasks and do not consider augmentations or human video.
A recently emerging line of works \cite{Palo2024KeypointAT, R+X} combines keypoint correspondence from vision models with the in-context learning capabilities of LLMs to transfer policies from human video after retrieval.
Such MLLM policies can only be deployed open-loop, due to the costly nature of MLLM inference, and hence are not reactive to scene changes.
Further, as MLLMs have not been trained with spatial data, they struggle to generalize to novel spatial arrangements and require dense scene coverage to obtain useful in-context demos. 
Closer to our work, recent works \cite{P3PO,PointPolicy,liu2025egozerorobotlearningsmart} parse keypoints from human demo videos and train keypoint-conditioned closed-loop policies, operating on point tracks predicted by an off-the-shelf tracker \cite{CoTracker}.
All above works require many videos to train generalizable policies and have mostly not been deployed for bimanual manipulation.
In our work, we train keypoint-conditioned policies on spatial augmentations performed on a single human demo video, offering sample-efficient and generalizable bimanual policies without need for teleoperation or extensive human videos.

\noindent \textbf{Demonstration augmentation} A recent line of works attempts to generate demonstrations from scratch using LLMs for task decomposition into sub-goals and motion primitives or reinforcement learning (RL) for completing the sub-goals \cite{Wang2023GenSimGR, Wang2023RoboGenTU, Hua2024GenSim2SR}.
However, the range and quality of the resulting data is often restricted by the capacity of the underlying planning
and RL modules.
The seminar work MimicGen \cite{MimicGen} and its follow-ups \cite{IntervenGen,SkillMimicGen,DexMimicGen} provide a more effective alternative.
Instead of generating data from scratch, MimicGen performs augmentations on human-collected demonstrations to adapt for novel spatial configurations,
by decomposing and re-synthesizing corresponding execution plans. 
% This approach
% is theoretically applicable to a wide range of manipulation
% skills and object types. 
DexMimicGen \cite{DexMimicGen} extends MimicGen’s strategy to support bimanual tasks and robot platforms.
However, plans produced by the MimicGen family \cite{MimicGen,SkillMimicGen,IntervenGen,DexMimicGen} are not
ready-to-use demonstration data in the form of state-action
pairs, and on-robot execution is necessary for collecting image observation data, required by the downstream policy. 
Therefore, they rely on rolling-out their execution plans on digital twin platforms, which bottlenecks data collection time while introducing additional sim-to-real
challenges for real-world deployment.
Closer to our work, DemoGen \cite{DemoGen} generates ready-to-use augmentations in a cost-effective manner.
As it considers 3D point-clouds as state representations for downstream policy learning, DemoGen supports obtaining the augmented states in a simulation-free fashion, simply via 3D rigid geometry. 
However, this approach only tackles spatial generalization, as point-cloud policies still struggle to generalize in novel object instances.
Further, the DemoGen framework only considers sequential single-arm tasks, which cannot be applied out-of-the-box for bimanual manipulation, while still requiring a few source demonstrations collected via teleoperation.
In our work, we extend DemoGen to handle bimanual manipulation tasks in a general framework, while considering keypoint states (obtained via parsing video) for downstream policy learning, thus also enabling category-level object generalization and robustness to background noise.

\section{METHOD}
\begin{figure}[!t]
  \centering
  \includegraphics[width=\textwidth]{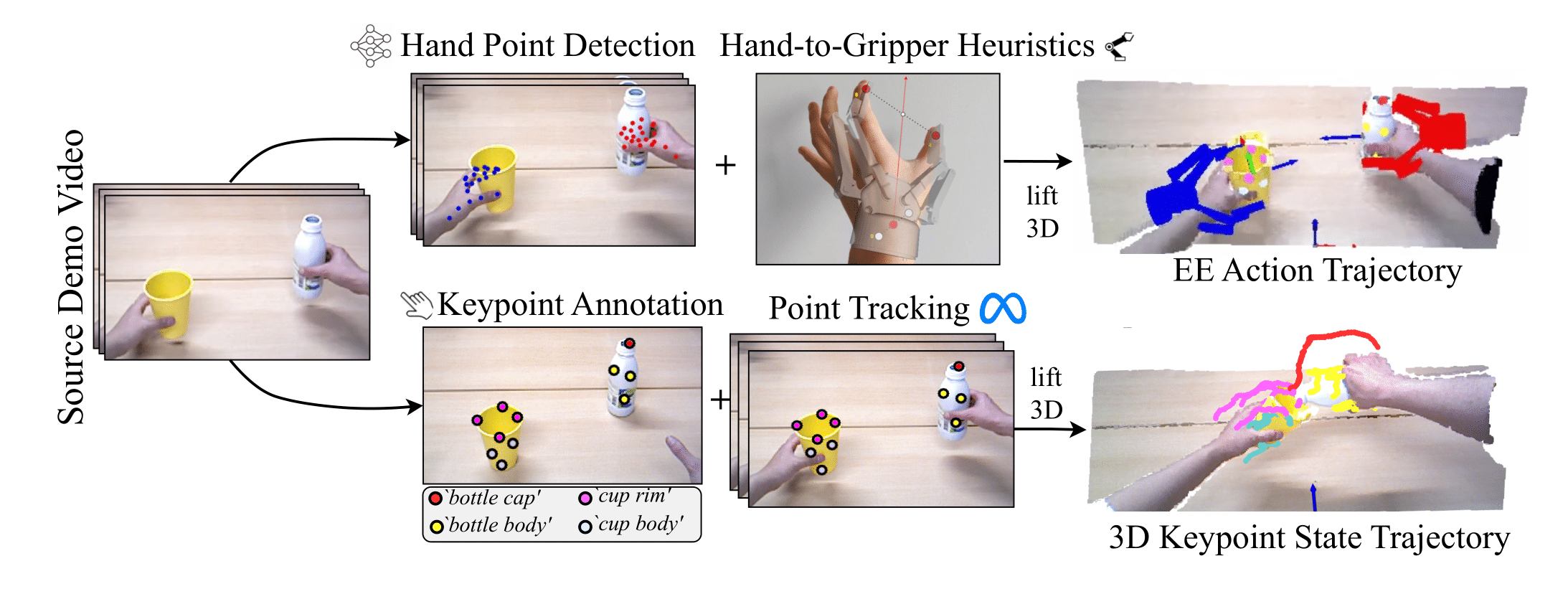}
  \caption{\textbf{Parse source video into state-action trajectory:} We use several off-the-shelf computer vision assets such as hand pose estimators and 2D point trackers, in-tandem with hand-to-gripper heuristics, to parse the source video into 3D keypoint and absolute end-effector action trajectories.}
  \label{fig:Parse}
    % \vspace{-4mm}
\end{figure}

Our problem setup states that given a single recorded video $V_M$, demonstrating an expert behavior for completing a bimanual task $M$, our goal is to obtain a policy $\pi_{M}$, trained solely from the source video, that can autonomously complete the task.
We explicitly seek generalization to novel spatial arrangements, object instances and background scene noise. 
In the following sections we describe our framework, \textit{PAD} (see schematic in Fig.~\ref{fig:fig1}), which decomposes the problem in three steps: (a) parsing the video into a robot-executable state-action trajectory (Sec.~\ref{method:parse}), (b) augmenting the trajectory at scale via bimanual TAMP (Sec.~\ref{method:augment}), and (c) distilling the generated data into a keypoint-conditioned diffusion policy (Sec.~\ref{method:distill}). 
% We also discuss implementation choices and extra steps in Sec.~\ref{method:implm}.

\noindent \textbf{Demonstration setup and assumptions} We record our demo video with an RGB-D camera, placed under the same orientation from the table as in our robot setup. We assume that a calibration procedure is performed, such that the extrinsics matrix of the recording camera relative to the tabletop matches the one of the robot. We call the result of this single camera-to-tabletop transform the \textit{task frame}, which we use to express all keypoint / action trajectories parsed from the video.
We also assume that the human demonstrator provides motion trajectories that lie within the robot's valid kinematic workspace.

\subsection{Parsing Video into Robot-Executable Data}
\label{method:parse}
The goal of this step is to parse the source RGB-D video $V_M = \left\{ (I_t, D_t) \right\}_{t=0}^{L-1}$, where $L$ the total number of frames (recorded at 30 fps), into the corresponding state-action transition trajectory $\zeta^{\texttt{demo}}_M = \left\{ (s_t,a_t) \right\}_{t=0}^{L-1}$.
In this work we define:   
$$s_t := \mathbf{P}_t^{kp} \; \in \mathbb{R}^{N \times 3}$$
$$a_t := (\mathbf{T}_t^{0}, g_t^{0}, \mathbf{T}_t^{1}, g_t^{1}) \; \in SE(3)^2 \times \{0,1\}^2$$
\noindent The state is represented by the 3D coordinates of $N$ prescribed keypoints, while the action as a pair of two end-effector (EE) poses $\mathbf{T}_t^{j} =(\mathbf{R}_{t}^j  |  \mathbf{x}_{t}^j) \in$ SE(3) and gripper open/close commands $g_t^{j} \in \{0,1\}$.
Here and for the rest of the manuscript we use the superscript $j \in \{0,1\}$ to denote each arm (e.g. 0-left and 1-right).
% Both poses and points are expressed wrt. the calibrated task frame.
% $a_t^{j} = [\mathbf{x}_t^{j}, \tilde{\mathbf{R}}_t^{j}, g_t^{j}]$, concatenating the end-effector's (EE) 3D position $\mathbf{x}_t^{j}$, 6D orthonormal rotation representation $\tilde{\mathbf{R}}_t^{j}$ [], as well as the gripper openness signal $g_t^{j} \in \{0,1\}$, for each of the two arms $j=0,1$.
% Both poses and keypoints are expressed with respect to the calibrated task frame.
% $a_t := (\mathbf{T}_{ee0}, \mathbf{T}_{ee1}) \in \mathbb{R}^{2 \times 4 \times 4 }$ as a pair of SE(3) poses for the two arms' end-effectors (EE), all expressed with respect to the calibrated task frame. 
The overall system for extracting keypoints and EE actions is illustrated in Fig.~\ref{fig:Parse}.

\noindent \textbf{Annotating and tracking keypoints} To obtain keypoint annotations, we adopt the method proposed by \cite{P3PO}. 
In particular, a user selects semantically meaningful 2D points on task-relevant objects in the first frame of the video $I_0$ through a GUI.
We also ask the user to provide a text label for groups of keypoints, corresponding to distinct object-parts that participate in contacts at different stages of the task (e.g. \textit{`bottle body'} for grasping the bottle and \textit{`bottle cap'} for pouring from it -- see bottom left of Fig.~\ref{fig:Parse}).
These will become useful downstream as positional embeddings for policy learning.
Next, we employ an off-the-shelf point tracker, Co-Tracker3 \cite{CoTracker}, to automatically track the initialized keypoints in the rest of the video frames $I_{t>0}$.
The depth maps $D_t$ are used to back-project the 2D keypoint coordinates to 3D.
To account for noise in sensor depth, we consider a 2D window around each back-projected pixel and select the median value after removing outliers.
Finally, we use the camera extrinsics to transform the keypoint tracks to task frame, resulting in the final 3D keypoint states $\mathbf{P}^{kp}_{0:L-1}$.

\noindent \textbf{Mapping hands to robot actions} Following previous works \cite{R+X}, we use HaMeR \cite{HaMeR}, a recent model trained for hand pose estimation in egocentric hand-object interaction videos \cite{Ego4D}.
Hand poses are parameterized through 21 points that correspond to the joints of the human hand according to the MANO model \cite{MANO:SIGGRAPHASIA:2017} (see top left of Fig.~\ref{fig:Parse}). 
We run HaMeR separately for each frame $I_t$ to obtain both 2D pixel and 3D point coordinates for each hand $\mathbf{p}_{t,k}^j, \, k=0,\dots,20$.
Depth maps $D_t$ are used to correct mis-calibration errors in HaMeR's 3D prediction, by matching a point which is clearly visible (e.g. wrist) with its back-projected depth and transferring the rest of the points such as their relative positions remain unchanged.
% Even though HaMeR directly provides 3D point coordinates, we find a discrepancy with the actual recorded data which accumulates over frames. 
% To tackle this, we use the depth maps $D_t$ to back-project the predicted 2D pixel coordinates to 3D, match a point which is always visible (e.g. wrist) with the HaMeR 3D prediction, and transfer the rest of the points such as their relative spatial relation remains unchanged. 
% This process provides a trajectory of 21 3D point coordinates for each hand: $\mathbf{p}^{j}_{1:L,k}$, with $k=0,...,20$.
To map hand points to robot poses, we use heuristics similar to the ones described in \cite{R+X}. 
In particular, of the 21 points, we use the tip of the index finger $\mathbf{p}^{j}_{t,\texttt{ind}}$, tip of the thumb $\mathbf{p}^{j}_{t,\texttt{th}}$ and the mid-wrist point $\mathbf{p}^{j}_{t,\texttt{wr}}$.
Then, we define the robot's EE position at the midpoint between the index and thumb tip and the rotation by aligning the line connecting the wrist and index-thumb midpoint with the gripper's approach direction (see top left of Fig.~\ref{fig:Parse}).
Formally, an EE pose $\mathbf{T}^{j}_t = (\mathbf{R}_{t}^j  |  \mathbf{x}_{t}^j)$ is given by:
\begin{equation*}
    \mathbf{x}_t^j = \tfrac{1}{2}(\mathbf{p}_{t, \texttt{th}}^j + \mathbf{p}_{t, \texttt{ind}}^j)
\end{equation*}
\begin{equation}
\mathbf{a}_t^j = \mathbf{p}_{t, \texttt{ind}}^j - \mathbf{p}_{t, \texttt{th}}^j, \;
\mathbf{b}_t^j = \mathbf{x}_t^j - \mathbf{p}_{t, \texttt{wr}}^j, \;
\mathbf{v}_t^j = \mathbf{a}_t^j \times \mathbf{b}_t^j \\
\end{equation}
\begin{equation*}
\mathbf{R}_t^j = \mathbf{I}_{3 \times 3} + [\hat{\mathbf{v}}_t^j]_{\times} + \frac{1 - \hat{\mathbf{a}}_t^j \cdot \hat{\mathbf{b}}_t^j}{\| \hat{\mathbf{v}}_t^j \|^2} \cdot [\hat{\mathbf{v}}_t^j]_{\times}^2
\end{equation*}

\noindent where $\times$ denotes cross-product, $\hat{\mathbf{v}}=\mathbf{v}/ \left\| \mathbf{v} \right\|$ unit-length normalization and $[\mathbf{v}]_{\times}$ the skew-symmetric matrix of $\mathbf{v}$ \cite{Sol2018AML}. 

To facilitate robot-executability, we use an Inverse Kinematics (IK) solver to identify "jumps" in the parsed action trajectory that lead to kinematic failures or collisions. We replace those actions with collision-free poses interpolated from the rest of the trajectory with cubic splines \cite{ReKep}. 
This process smoothens the action trajectory and ensures executability, given that most of the recorded trajectory is within the robot's kinematic range.

\subsection{Trajectory Augmentation via Bimanual Task-and-Motion Planning}
\label{method:augment}

The goal of this step is to generate a large-scale dataset $\mathcal{D}^{aug}_M = \{ \tilde{\zeta}^{(i)} \}$, where each trajectory $\tilde{\zeta}$ is generated by performing spatial augmentations over the source demo $\zeta^{\texttt{demo}}_M$.
Like previous works \cite{MimicGen,DemoGen}, the augmentations are orchestrated by a TAMP procedure, where the action trajectory is split into \textit{motion} and \textit{skill} segments.
Motion segments correspond to the robot moving in free space (e.g. approaching the bottle), while skill segments to robot-object or object-object interactions that manifest a manipulation skill (e.g. grasping the bottle, pouring from bottle to cup).
The main idea is to cover the entire workspace by moving the task-relevant objects around, and appropriately adapt the actions in each motion and skill segment through motion planners and SE(3)-equivariant transforms respectively.
In PAD we extend this methodology for bimanual tasks, taking special care of related issues that arise with arm collaboration.
In the rest of the section we provide step-by-step details of our augmentation framework.

\begin{figure}[!t]
  \centering
  \includegraphics[width=\textwidth]{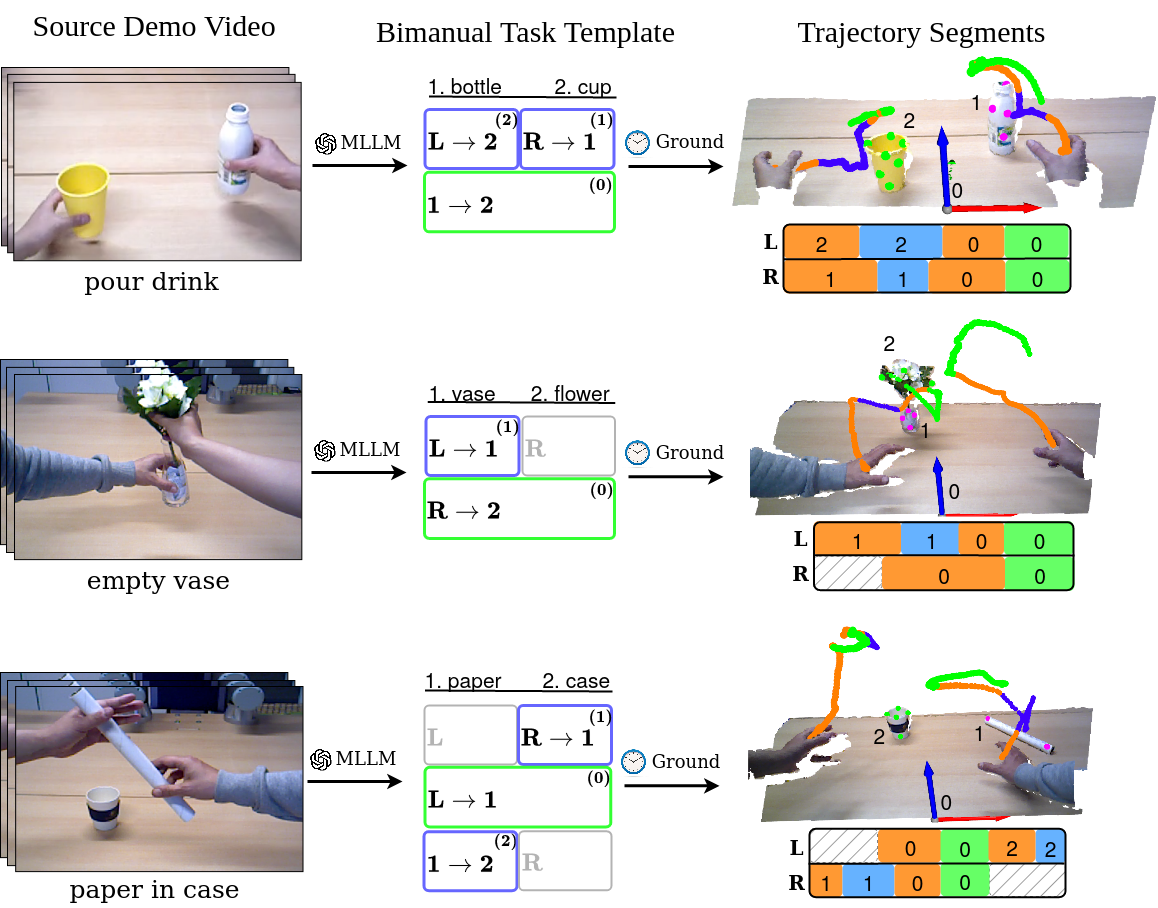}
  \caption{\textbf{Grounding trajectory segments with bimanual task templates.} The video (\textit{left}) is abstracted into a template (\textit{middle}), denoting task stages (as blocks), given by contacts (arrows), related reference frames (number in parenthesis) and arm synchronization (\textcolor{green}{green} block) or not (\textcolor{blue}{blue} block). Timestamp segments for each arm-stage are grounded in the demo trajectory, where \textcolor{orange}{orange} blocks correspond to \textit{motion}, \textcolor{gray}{gray} blocks to \textit{idle}, \textcolor{blue}{blue} blocks to \textit{asynchronous skill} and \textcolor{green}{green} to \textit{synchronous skill} segments.}
  \label{fig:Augm1}
    % \vspace{-4mm}
\end{figure}

\noindent \textbf{Bimanual task templates} The demonstrated bimamual manipulation task $M$ is abstracted into a symbolic template $\Pi_M$, which contains: (a) a set of integers $\{1,...,K\}$, each corresponding to a unique task-relevant object in the scene, and (b) a list of high-level actions, each represented by a \textit{contact}, i.e. a pair sampled from the set $\left\{ ee^0, ee^1 \right\} \cup \left\{ 1,\dots,K \right\}$ indicating a robot-object or object-object contact, as well as a \textit{reference} index $k \in [0,K]$, indicating whether the action depends on some object's local reference frame ($k>0$) or not ($k=0$).
Each element in the list can include two actions (one for each arm), indicating parallel execution of two asynchronous motions, or a single bimanually-synchronized action.
% An action with non-receiving robot contact indicates the arm remaining idle.
Examples of task templates are illustrated in Fig.~\ref{fig:Augm1}.

The task templates serve as essential information for grounding segments in the subsequent TAMP procedure, by means of: (a) decomposing the task into distinct object-centric stages, (b) declaring hand-object assignments, (c) assigning related reference frames to each stage-arm pair, and (d) declaring stages that require bimanual synchronization.
The task templates are general, i.e. they don't depend on the semantics of the appearing objects and dont require skill labels.
Instead, they provide a high-level overview of the task in terms of what contacts with / between objects take place, in what order, and whether the arms need to be synchronized.
Further, they are composable, as arbitrary long-horizon tasks can be constructed by chaining templates for specific short-horizon skills.
In practice, we obtain $\Pi_M$ by feeding sub-sampled frames of the video with a few in-context examples to a MLLM \cite{GPT4}, akin to \cite{R+X}.

\noindent \textbf{Grounding timestamp segments} Our next step is to ground each stage of our task template into a sequence of motion and skill segments for each arm, with precise timestamps in the parsed action trajectory. 
First, we use GroundedSAM \cite{GroundedSAM} to segment the $K$ task-relevant objects in the initial video frame $I_0$, and back-project each region to 3D to obtain a masked object point-cloud. 
For each object $o_k$, we calculate its point center and place a local reference frame there $\mathbf{T}^{o_k} \in $ SE(3), resulting in the \textit{initial object configuration} $\mathcal{O}= [\mathbf{T}^{o_1}, \dots, \mathbf{T}^{o_K}]$.
The reference annotation $k$ in arm-stage pairs of $\Pi_M$ can then be identified with the corresponding reference frame $\mathcal{F}_i^j \in \left\{ \mathbf{T}^{task} \right\} \cup \mathcal{O}$, which we call the \textit{assigned} frame for arm $j$ in stage $i=1,\dots,n$.
% Here, the identity matrix is identified with stage-arm pairs that are annotated with reference $k=0$, hence are assigned to the global task frame (see Fig.X).
Here, the global task frame $\mathbf{T}^{task}$ is assigned to stage-arm pairs that are annotated with reference $k=0$.
Next, for each stage, we classify demo timestamps into motion or skill segments. 
Following \cite{DemoGen}, we classify \textcolor{blue}{asynchronous skill} segments, assigned to a frame $\mathbf{\mathcal{F}}_i^j$, by calculating the distance between the frame's center and the assigned arm's EE position, and checking whether it falls within a specified threshold.
Similarly, to adapt to syncrhonous stages, we classify \textcolor{green}{synchronous skill} segments as the timestamps whose distance between the two EE poses falls within a specified threshold.
Synchronous segments will always have the exact timestamps between arms.
The intermediate timestamps between two skill segments are classified as \textcolor{orange}{motion} segments.

Formally, for a timestamp segment
$
    \tau = [t_{a},t_{b}] \subseteq [0,L),
$
we use the bracket notation $[\cdot]$ to denote slicing \cite{DemoGen}, such that $\zeta[\tau]=(s_{t_a:t_b}, a_{t_a:t_b})$.
% $u[\tau]=u_{t=t_{a}:t=t_{b}}$ for any time-dependent quantity $u_t$
Then, the state-action trajectory wrt. the initial configuration $\mathcal{O}$ can be written as \footnote{The skill segments are randomly colorized for illustration purposes, in general skill segments can arbitrarily vary between synchronous/asynchronous throughout stages.}:
% \begin{equation*}
%     A^j \parallel \mathcal{O} = \left[ \textcolor{orange}{a^j[\tau^{m}_1]}, \,  \textcolor{blue}{a^j [\tau_1^{s}]}, \dots,  \textcolor{orange}{a^j [\tau^{m}_{n}]}, \,  \textcolor{green}{a^j [\tau_n^{s}]} \right]
% \end{equation*}
% \noindent or also including states by considering $\zeta[\tau]=(s_{t_a:t_b}, a_{t_a:t_b})$:
\begin{equation*}
    Z \parallel \mathcal{O} = \left[ \textcolor{orange}{\zeta[\tau^{m}_1]}, \,  \textcolor{blue}{\zeta [\tau_1^{s}]}, \dots,  \textcolor{orange}{\zeta [\tau^{m}_{n}]}, \,  \textcolor{green}{\zeta [\tau_n^{s}]} \right]
\end{equation*}
\noindent where superscripts $\{m,s \}$ denote a motion and skill segment respectively. 
% During object-assigned skill segments where grasping happens (i.e. $\Delta g^j_t>0$) we also record the relative pose between the EE and the assigned object's frame $\mathbf{\Delta T}^{o_k}_{ee} = \mathbf{T}_{o_k}^{-1} \cdot \mathbf{T}_{ee}  $ at that timestep, which is necessary later for augmenting keypoints.

\begin{figure}[!t]
  \centering
  \includegraphics[width=\textwidth]{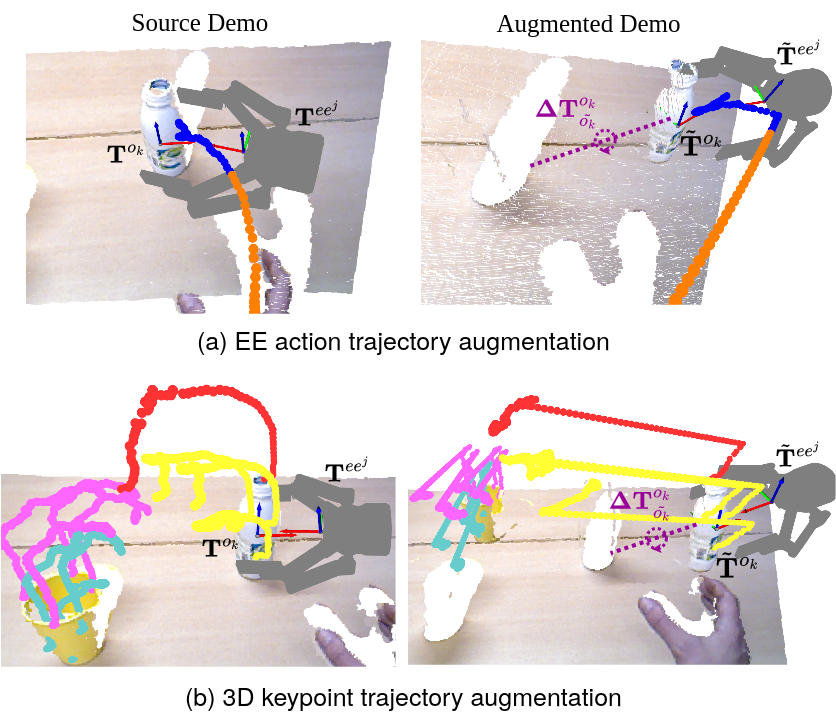}
  \caption{\textbf{Augmenting state-action trajectories with SE(3)-equivariant
transforms.} \textit{(Top)} The EE poses are transformed according to the new
object pose such as their relative pose remains invariant for the entire skill
segment (in blue color). \textit{(Bottom)} We assume the new keypoints will move rigidly
with the EE pose, according to their relative pose during grasping.}
  \label{fig:Augm2}
    \vspace{-2mm}
\end{figure}

With timestamp segments in place, we vary the initial object configuration $\tilde{\mathcal{O}} = [\mathbf{T}^{\tilde{o_1}}, \dots, \mathbf{T}^{\tilde{o_K}}]$ with SE(3) transformations and obtain
$
    \tilde{\zeta} := \tilde{Z} \parallel \tilde{\mathcal{O}} 
$
, where $\tilde{\zeta}$ an augmented demo that has the same sequence of motion-skill segments as $\zeta^{\texttt{demo}}_M$, but the state-actions in each segment are augmented according to $\tilde{\mathcal{O}}$, as described below (see also Fig.~\ref{fig:Augm2}).

\noindent \textbf{Action augmentations} Let $\mathbf{\Delta T}_{\tilde{o_k}}^{o_k}=(\mathbf{T}^{o_k})^{-1} \cdot \mathbf{T}^{\tilde{o_k}}$ the relative transformation between the demo and augmented configuration of object $o_k$.
Then, the augmented action trajectory for arm $j$ corresponding to a skill segment $\tau^{s}_i$ with assigned object frame $\mathcal{F}_i^j=\mathbf{T}^{o_k}$, is given by:
\begin{equation}
    \tilde{\mathbf{T}}^j[\tau_i^s] = \mathbf{T}^j[\tau_i^s] \cdot  \mathbf{\Delta T}_{\tilde{o_k}}^{o_k},
\end{equation}
$$
\tilde{g}^j[\tau_i^s]=g^j[\tau_i^s]
$$
which is an SE(3)-equivariant transform that preserves the relative pose between object and robot throughout the segment: $\mathbf{\Delta T}_{ee^j}^{o_k}[\tau_i^s] = \mathbf{\Delta T}_{\tilde{ee}^j}^{\tilde{o_k}}[\tau_i^s]$. This augmentation maintains an equal number of steps as in the demo skill segment and is the same for both synchronous and asynchronous segments.
Notice that in cases of skill segments where there is no assigned object (i.e. $\mathcal{F}_i^j=\mathbf{T}^{task}$), it's $\mathbf{\Delta T}_{task}^{task}=\mathbf{I}_{4x4}$, so the equation yields $\tilde{\mathbf{T}}_t^j = \mathbf{T}_t^j $, i.e. the segment's trajectory is exactly replayed as in the demo, as it's independent of any object that has been transformed.
In all cases the gripper signal remains unchanged, since it is invariant to the object transforms.
For motion segments, the goal is to connect the last pose of the previous skill segment with the first pose of the current one, which is obtained via motion planning \footnote{For uncluttered scenarios, linear interpolation suffices. Alternatively, third-party motion planners [26] for obstacle avoidance should be employed.}
:
\begin{equation}
    \tilde{\mathbf{T}}^j[\tau_i^m] = \texttt{MotionPlan} \left( \tilde{\mathbf{T}}^j_{t=\tau_{i-1}^s[-1]}, \, \tilde{\mathbf{T}}^j_{t=\tau_{i}^s[0]}  \right)
\end{equation}
$$
\tilde{g}^j[\tau_i^m]=g^j_{t=\tau_{i}^m[0]}
$$
\noindent with the demo EE poses at $t=0$ used as the start pose for the first stage.
The gripper signal is copied from the start timestep to the entire segment, as there are no contacts taking place during motion segments.
The full augmented action trajectory under the new object configuration is obtained by concatenating all segments:
\begin{equation}
\label{eqn4}
    \tilde{A} \parallel \tilde{\mathcal{O}} = \left[ \tilde{a}[\tau^{m}_1], \, \tilde{a} [\tau_1^{s}],  \dots, \tilde{a}[\tau^{m}_n], \, \tilde{a} [\tau_n^{s}] \right]
\end{equation}
\noindent where $\tilde{a}[\tau_i]= \left( \tilde{\mathbf{T}}^0[\tau_i], \tilde{g}^0[\tau_i], \tilde{\mathbf{T}}^1[\tau_i], \tilde{g}^1[\tau_i] \right)$ the augmented bimanual action trajectory for each segment.

\noindent \textbf{Keypoint augmentations} We make the 3D rigidity assumption \cite{DemoGen} to augment the keypoint trajectory under the new object configuration and augmented actions. 
In particular, throughout skill segments $\tau_{1:n}^s$, we keep track of timestamps where grasping $t_g$ ($\Delta g^j_t>0$) or releasing $t_r$ ($\Delta g^j_t<0$) actions took place.
For intermediate steps between grasp and release, we consider the object to be attached to the assigned EE and that it moves rigidly with it in 3D space, i.e. their relative pose remains invariant.
If $\tilde{\mathbf{T}}^j_{t_g}$ the augmented EE pose during attachment of arm $j$ with object $o_k$ and $\mathcal{P}^{o_k}_t$ the homogeneous coordinates of all $N_k$ keypoints that belong to that object, then the augmented keypoints in a future step $t$ within successive $[t_g,t_r]$ intervals are given by:
\begin{equation}
\tilde{\mathcal{P}}^{o_k}_t = \tilde{\mathbf{T}}^j_t \cdot (\tilde{\mathbf{T}}_{t_g}^j)^{-1} \cdot  \tilde{\mathcal{P}}^{o_k}_{t_g}
\end{equation}
Between release and grasp intervals $[t_r,t_g]$, the keypoints are kept static since there is no attachment: $\tilde{\mathcal{P}}^{o_k}_t=\tilde{\mathcal{P}}^{o_k}_{t_r}$, with $ \tilde{\mathcal{P}}^{o_k}_{0} = \mathbf{\Delta T}_{\tilde{o_k}}^{o_k} \cdot \mathcal{P}^{o_k}_0$ for the start of the episode $t=0$.
Coordinates $\tilde{\mathcal{P}}_t^{o_k} \in \mathbb{R}^{4 \times N_k} $ are converted back to non-homogeneous $\tilde{\mathbf{P}}_t^{o_k} \in \mathbb{R}^{N_k \times 3}$ and concatenated across all objects' keypoints: $\tilde{s}_t = [\tilde{\mathbf{P}}^{o_1}_t, \dots, \tilde{\mathbf{P}}^{o_K}_t]=\tilde{\mathbf{P}}_t^{kp}$.
Applying this to all segments we obtain the augmented state trajectory:
\begin{equation}
\label{eqn5}
    \tilde{S} \parallel \tilde{\mathcal{O}} = \left[ \tilde{s}[\tau^{m}_1], \, \tilde{s} [\tau_1^{s}],  \dots, \tilde{s}[\tau^{m}_n], \, \tilde{s} [\tau_n^{s}] \right]
\end{equation}
and finally the entire augmented state-action trajectory by combining equations (4) and (6):
\begin{equation}
    \tilde{Z} \parallel \tilde{\mathcal{O}} = \left[ (\tilde{s}[\tau^{m}_1],\tilde{a}[\tau^{m}_1]), \dots, (\tilde{s}[\tau^{s}_n], \, \tilde{a} [\tau_n^{s}]) \right]
\end{equation}

\begin{figure}[!t]
  \centering
  \includegraphics[width=\textwidth]{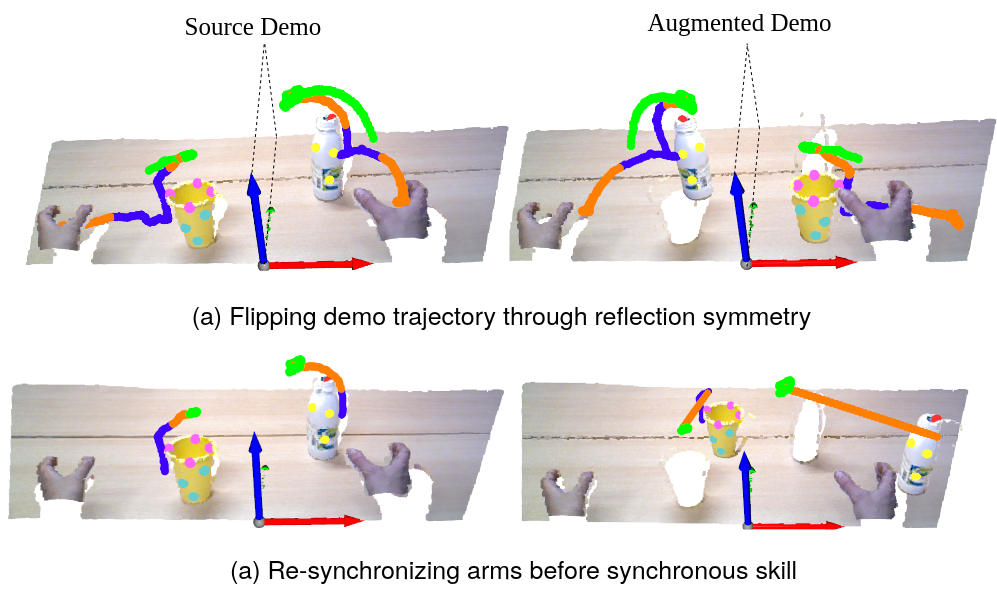}
  \caption{\textbf{Handling bimanual issues during augmentation.} \textit{(Top)} We generate novel hand-object assignment and augment the trajectory by reflecting according to the principal symmetry plane (YZ plane in the image). \textit{(Bottom)} The number of motion planning steps for each arm is inferred in the augmented demo such that they smoothly end up at the same place before the synchronous stage.}
  \label{fig:Augm3}
    \vspace{-2mm}
\end{figure}

\noindent \textbf{Handling bimanual issues} Unlike previous works \cite{DexMimicGen}, which assume the same number of steps between demo-augmented motion segments (and hence small perturbations in the objects' augmented poses), in this work we deal with the case where objects have been moved significantly further than their original configuration, resulting in unfeasible motion plans within the demo's number of steps.
We use a constant velocity to infer the number of steps for each motion segment based on the covered distance between start and goal pose.
However, this process naturally breaks the two arms' synchronization. 
To tackle this, we compare the start timestamps of the two arms' augmented trajectories during synchronous skill segments, and repeat the last pose to the arm that is the "earliest" for the remaining deficit. This process ensures that the two arms will be in the same relative pose as in the demo when a synchronous segment starts (see Fig.~\ref{fig:Augm3}-b).
Further, some augmentations will move objects outside the kinematic range of their assigned arm. To deal with this, we constraint the transforms to only spawn within-range final object configurations, and separately augment the initial configuration to deal with novel hand-object assignments.
In particular, since we have defined our task frame to be symmetric wrt. the two arms, we can obtain a mirrored version of the demo by simply reflecting keypoints and EE poses wrt. the principal plane of symmetry (see Fig.~\ref{fig:Augm3}-a). The reflected demo and initial configuration is fed to the same TAMP procedure to generate many augmentations for the mirrored case.

Overall, our augmentation framework entertains several desirable properties: (a) it is \textbf{general}, as it can represent a broad variety of bimanual manipulation tasks, (b) it is \textbf{simulation-free}, i.e. it doesn't require simulation rollouts with digital twins and hence significantly boosts collection time ($10^3$ demos in under 1 min.), (c) it is \textbf{view-invariant}, as all augmentation data are generated wrt. a calibrated task frame, independent of downstream robot camera placement, (d) it is \textbf{bimanual-aware}, as it handles common issues of previous bimanual demo augmentation frameworks, such as out-of-range hand-object assignments and de-synchronization, and (e) it is \textbf{embodiment-agnostic}, as all planning is done in EE space, and any off-the-shelf motion planner can be plugged into the described TAMP framework, appropriate for a given robot morphology.

\subsection{Keypoint-conditioned Bimanual Visuomotor Policy Learning}
\label{method:distill}
% $s_t := \mathbf{P}^{}_{t-T_o+1 \, : \, t} \in \mathbb{R}^{N \times T_o \times 3}$
The goal of this step is to distill the generated dataset $\mathcal{D}_M^{aug}$ into an autonomous policy $\pi_{M}$.
To facilitate learning keypoint-conditioned policies, we adapt RDT \cite{RDT}, a recent diffusion transformer model based on DiT \cite{Peebles2022ScalableDM}, employed for bimanual visuomotor policy learning.
Formally, the policy samples from the distribution $p(\mathbf{A}_t | \mathbf{o}_t)$, where $\mathbf{A}_t=\mathbf{a}_{t:t+H-1}$ an action chunk \cite{zhao2023learningfinegrainedbimanualmanipulation} over a prediction horizon $H$, and $\mathbf{o}_t=(\mathbf{q}_t, \mathbf{P}_{t-T_o+1:t}^{kp})$, where $\mathbf{q}_t$ the current proprioception state and $\mathbf{P}_{t-T_o+1:t}^{kp} \in \mathbb{R}^{N \times T_o \times 3}$ the 3D keypoint coordinates over an observation window of $T_o$.
We use absolute EE control \cite{DemoGen}, and parameterize actions as $\mathbf{a}_t \in \mathbb{R}^{20}$, where we flatten each EE's 3D position, 6D rotation vector (first two columns of rotation matrix) \cite{KALM} and gripper open/close command into a single vector representation.

\begin{figure}[!t]
  \centering
  \includegraphics[width=\textwidth]{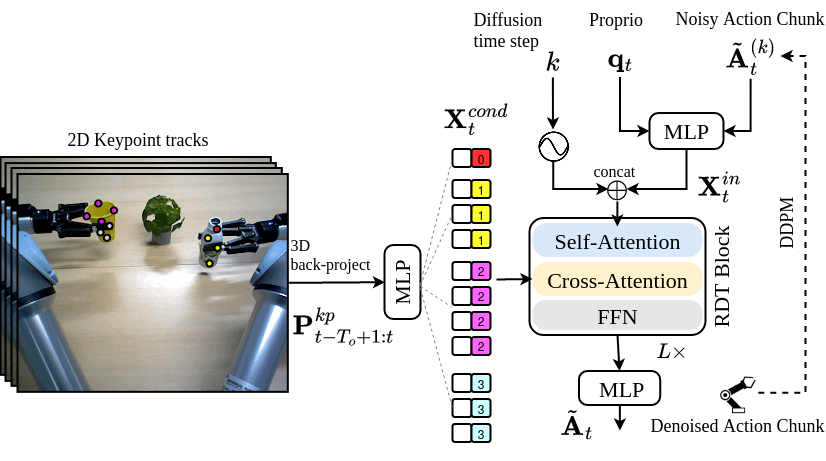}
  \caption{\textbf{Kp-RDT architecture overview.} We adapt RDT to receive 3D keypoint track conditions. We add learnable embeddings to each keypoint's token to discriminate between distinct keypoint groups (i.e. \textit{group embeddings}), without needing a fixed order for keypoints within a group.}
  \label{fig:KpRDT}
    \vspace{-3mm}
\end{figure}

\noindent \textbf{Kp-RDT architecture.} Our efforts focus on adapting the conditional branch of the cross-attention layers in RDT to keypoint track states, which we term \textit{Kp-RDT} (see Fig.~\ref{fig:KpRDT}).
We flatten the $(T_o,3)$ dimensions and encode each keypoint's observation with an MLP, resulting in a token sequence $\mathbf{X}^{cond}_t \in \mathbb{R}^{N \times C}$, where $C$ the transformer's hidden size.
Here we omit positional encodings on the keypoint token sequence, which drops the requirement that keypoints will be matched in a fixed order.
Instead, we use the keypoint group annotations (see Sec.~\ref{method:parse}) to encode identifying information about each keypoint via \textit{group embeddings}, which are added to the keypoint token sequence after mapping all keypoints to their unique group indices, embedded in $\mathbb{R}^C$ as learnable parameters.
The self-attention branch is as in RDT, where the diffusion step $k$ is encoded with a sinusoid MLP encoder \cite{RDT}, while proprioception $\mathbf{q}_t$ and noisy action chunk $\tilde{\mathbf{A}}_t^{(k)}$ with the same MLP encoder, resulting in the input token sequence $\mathbf{X}^{in}_t \in \mathbb{R}^{(2+H) \times C}$.
After successive application of $L$ Kp-RDT layers, the last $H$ transformed tokens are mapped to denoised action chunks $\tilde{\mathbf{A}}_t^{}=f_{\theta}(\mathbf{o}_t,\tilde{\mathbf{A}}_t^{(k)},k)$ with an MLP decoder. 

\noindent \textbf{Training \& Inference.} Following standard practices \cite{RDT}, we train the parameters $\mathbf{\theta}$ of Kp-RDT $f_{\mathbf{\theta}}$ to estimate a clean action chunk from a noisy one:
% $\tilde{\mathbf{A}}_t^{(k)}=\sqrt{\bar{\alpha}^{(k)}}\mathbf{A}_t^{}+\sqrt{1-\bar{\alpha}^{(k)}}\mathbf{\epsilon}$:
\begin{equation}
\mathcal{L}({\theta}):=\mathcal{L}\left( \mathbf{A}_t^{}, \; f_{\theta}(\mathbf{o}_t, \sqrt{\bar{\alpha}^{(k)}}\mathbf{A}_t^{}+\sqrt{1-\bar{\alpha}^{(k)}}\mathbf{\epsilon},k) \right)    
\end{equation}
where $k \sim \mathcal{U}(1,K)$ the diffusion step, $\mathbf{\epsilon} \sim \mathcal{N}(\mathbf{0}, \mathbf{I})$ random noise, $(\mathbf{o}_t, \mathbf{A}_t) \sim \mathcal{D}^{aug}_M$ and $\bar{\alpha}^{(k)}$ coefficients pre-defined by a DDPM scheduler \cite{Nichol2021ImprovedDD}.
% \begin{equation}
%     \mathbf{A}^{(k-1)}_t=\frac{\sqrt{\bar{\alpha}^{(k-1)}}\beta^{(k)}}{1-\bar{\alpha}^{(k)}}\mathbf{A}_t^{(0)}+\frac{\sqrt{\alpha^{(k)}}(1-\bar{\alpha}^{(k-1)})}{1-\bar{\alpha}^{(k)}}\mathbf{A}_t^{(k)}+\sigma^{(k)}\mathbf{z}
% \end{equation}
% We sample state-action pairs $(\mathbf{o}_t, \mathbf{A}_t)$ from our dataset $D^{aug}$ at fixed control frequency of 10 Hz. 
We train using separate L1 losses and noise schedulers for position and rotation logits and binary cross-entropy losses for the gripper logits \cite{KALM}.
For inference, we sample a clean action chunk by iteratively denoising  for a small number of diffusion steps (e.g. K=10), starting from $\tilde{\mathbf{A}}_t^{(K)} \sim \mathcal{N}(\mathbf{0}, \mathbf{I})$ \cite{RDT}.

\begin{figure}[!t]
  \centering
  \includegraphics[width=\textwidth]{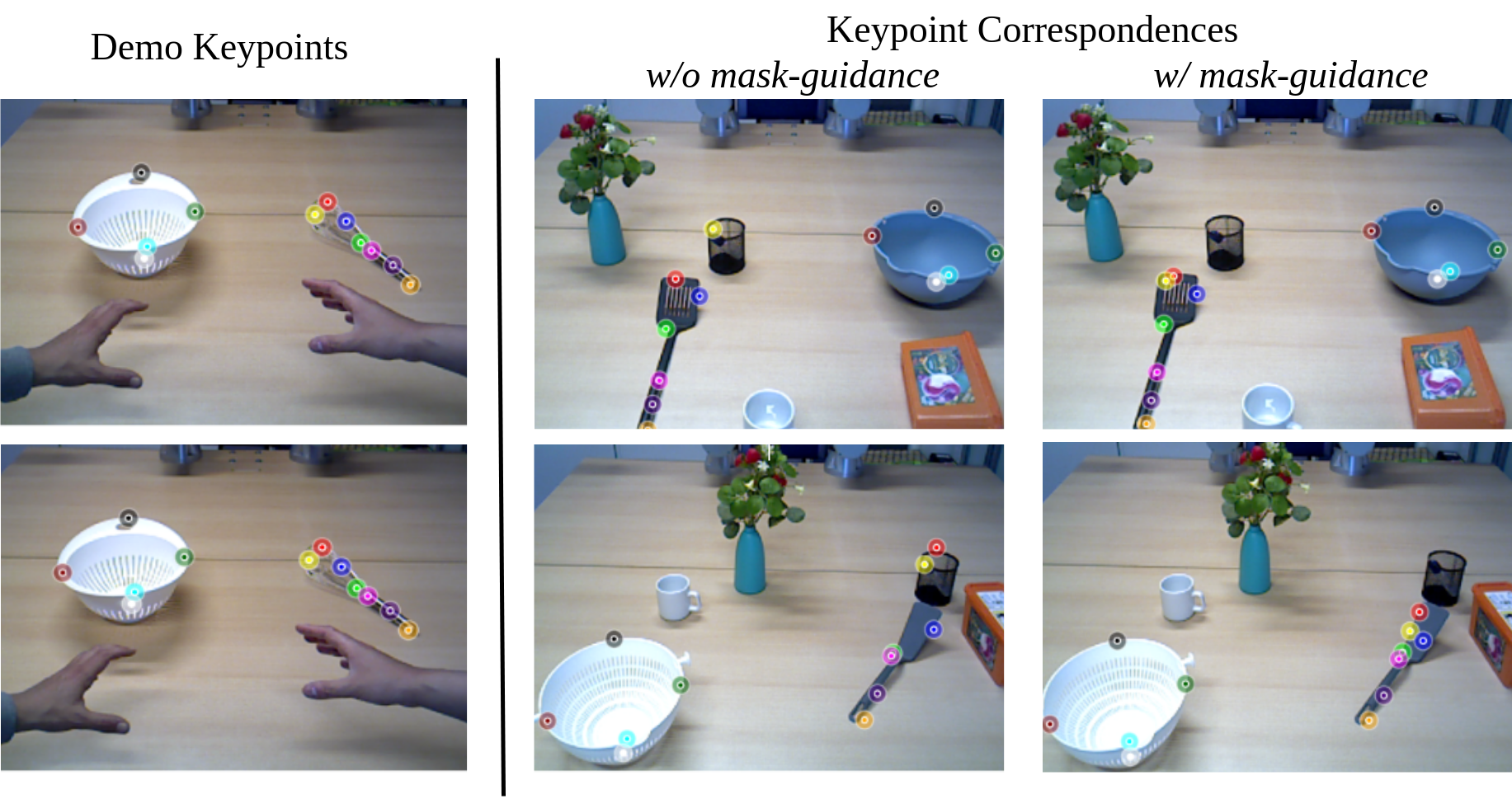}
  \caption{\textbf{Mask-guided keypoint correspondence} improves localization in scenes with background distractors, by constraining the outputs in the corresponding object's segmentation masks.}
  \label{fig:Corr}
    % \vspace{-4mm}
\end{figure}

To obtain keypoint states during inference, we first run keypoint correspondence with SD-DINOv2 \cite{SDDINOv2} on the initial RGB observation to initialize keypoint coordinates, and then employ Co-Tracker3 \cite{CoTracker}, adapted for on-the-fly execution \cite{PointPolicy}, to track the 2D coordinates in successive frames at real-time. 
Pixel coordinates are then back-projected to 3D and transformed to task frame to obtain $\mathbf{P}^{kp}_{t-T_o+1:t}$.
Importantly, we extend the correspondence algorithm to be \textit{mask-guided}, by first running GroundedSAM \cite{GroundedSAM} to obtain pixel-wise masks of the task-relevant objects in the observed image, and then push cosine similarity between demo-observed SD-DINOv2 features to zero for areas outside the masks.
We find this adaptation to significantly improve correspondence results, especially in presence of distractors (see Fig.~\ref{fig:Corr}).

\noindent \textbf{Implementation Details.} We sample state-action pairs during training at a fixed control frequency of 10 Hz.
To emulate keypoint tracking errors and partial visibility, we add Gaussian noise to the keypoint coordinates and randomly dropout a subset of them during training.
In practice, we mostly use an observation history of $T_o=8$, which we find to give robust states for the policy, and an action chunk horizon of $H=16$. 
We then execute only $T_a=4$ actions (i.e. $ 0.4$sec in 10 fps), akin to the \textit{receding horizon planning} strategy \cite{DiffusionP}.
In cases of non egocentric views (see Sec.~\ref{exp:sim}), we observed that replacing the point tracker with the keypoint forward model based on the 3D rigidity assumption described in Sec.~\ref{method:augment} can provide more robust tracking. 
Finally, our training setup can be easily adapted to obtain \textit{trajectory-level diffusers} \cite{KALM,DiffuserActor}, which predict the entire action trajectory from the initial keypoint state observation using a fixed number of steps, e.g. $H=96$, and execute it open-loop with an IK motion planner.
The latter implementation doesn't require online point-tracking, since it operates only on the initial observation, but the resulting policy loses its reactive behavior since it is run open-loop.

% \subsection{Implementation Deteails}
% \label{method:implm}

\section{EXPERIMENTS}
We design our experimental setup to answer the following questions:
(a) How does our augmentation framework compare to state-of-the-art bimanual demo augmentation approaches that use simulation rollouts? (b) What are the benefits in terms of task performance, data collection time and sample-efficiency? (c) Can our single video-trained policies generalize across spatial arrangements, object instances and background noise in real-world tasks? and (d) How does our proposed Kp-RDT compare to other recent keypoint-based policies in terms of success-efficiency trade-off?

We explore the first two questions by evaluating on 4 bimanual tasks of the DexMimicGen simulator \cite{DexMimicGen}, where we compare our approach using the generated data released by the authors. We explore the latter two questions by evaluating PAD on 6 diverse bimanual tasks recorded in a single video and executed with a real robot, focusing on a broad range of object instances to investigate generalization.

\begin{figure}[!t]
  \centering
  \includegraphics[width=\textwidth]{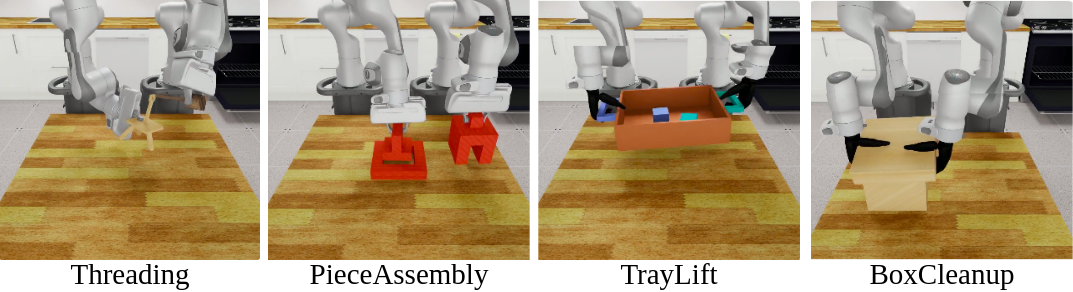}
  \caption{\textbf{Test tasks in DexMimicGen \cite{DexMimicGen} benchmark.} We evaluate on 4 bimanual Panda tasks in DexMimicGen with randomized object placements, using either parallel jaw or dexterous grippers.}
  \label{fig:dexmimicgen}
    \vspace{-2mm}
\end{figure}

\subsection{Simulation Experiments}
\label{exp:sim}

We conduct experiments on 4 bimanual tasks from the DexMimicGen benchmark \cite{DexMimicGen}, which is based on the robosuite framework \cite{MimicGen} that utilizes the MuJoco simulator.
We focus on the bimanual Panda tasks, two with parallel jaw gripper and two with dexterous arm (see Fig.~\ref{fig:dexmimicgen}).
We refer the reader to the DexMimicGen paper \cite{DexMimicGen} for details on the tasks. 
The focus of this section is to compare the augmentation pipelines between DexMimicGen and our work, in terms of final success rate, data collection time and sample-efficiency.
To that end, we use the 1000 demos per task released by the authors and replay them offline to save higher-resolution RGB-D observations.
We then pick a random demo (with three seeds) and use our augmentation framework to generate our own training dataset.

\noindent \textbf{Baselines} We use the re-collected DexMimicGen data to train three different kinds of image policies: (a) RGB (scratch), where we use the RGB observation encoded with a ResNet-18, (b) RGB-D (scratch), where we encode both RGB and depth separately with a ResNet-18 and concatenate to obtain an observation, and (c) RGB (SD-DINOv2), where we feed the RGB image to SD-DINOv2 \cite{SDDINOv2} (similar to our method) to obtain feature maps as condition to the policy.
For fair comparisons with our work, we use single agent-view observation and exactly the same RDT backbone \cite{RDT}, with only difference the type of conditioning used (image vs. keypoint).
Kp-RDT is trained with our own augmentation dataset and the image policies with DexMimicGen demos, while all methods are evaluated on the same $1000$ test scenes generated for each of the three source demo seeds.
For tracking in Kp-RDT, we use the 3D keypoint forward model based on the 3D rigidity assumption instead of Co-Tracker, since the provided view is non egocentric and most of the objects are occluded by the robot during execution.

\begin{figure}[!t]
  \centering
  \includegraphics[width=\textwidth]{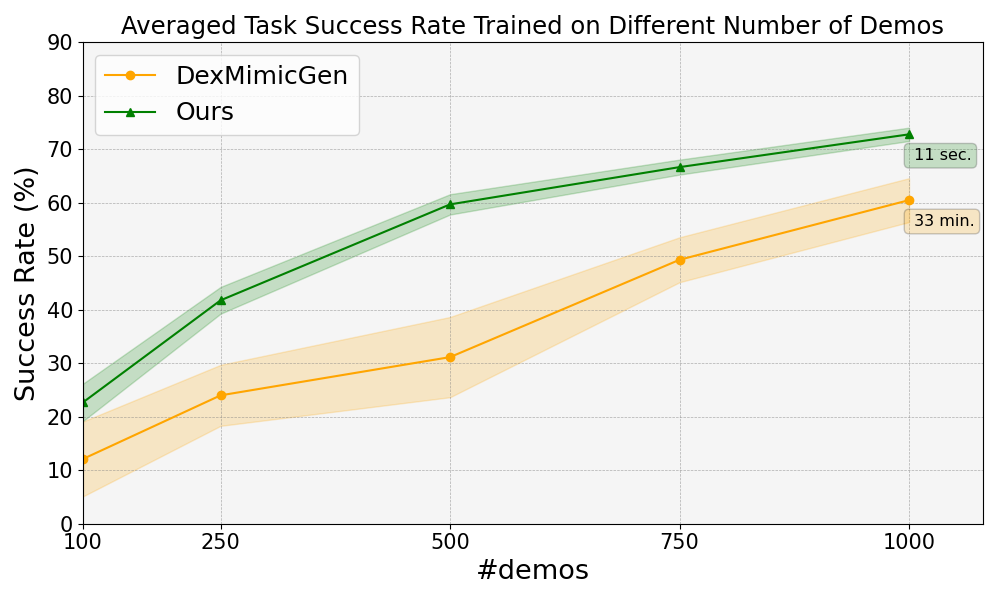}
  \caption{\textbf{Sample-efficiency analysis}. PAD leads to higher success rates and faster learning compared to DexMimicGen due to keypoint abstractions, while doing so by generating data in a much more efficient manner. See text for details on collection time computation. Mean and std results are averaged over 3 source demo seeds.}
  \label{fig:VsDexMimicGen}
    \vspace{-2mm}
\end{figure}

\noindent \textbf{Sample and cost efficiency} We compare performance for different number of augmented demos between the best image policy and our Kp-RDT.
We run experiments for the \textit{Threading} and \textit{PieceAssembly} task, and report averaged success rates in Fig.~\ref{fig:VsDexMimicGen}.
To compute collection time, we use the same method as described in \cite{DemoGen}.
In particular, both for our method and for DexMimicGen, we don't consider multi-processing / GPU parallelization and simply add the total time for a single process.
For DexMimicGen, we multiply the average time taken in our hardware to rollout one demo with the number of demos used.
Since our method doesn't need simulation rollouts, it is significantly faster, achieving generation of $10^3$ demos within $11$sec, compared to $33$min. by DexMimicGen.
Further, our Kp-RDT policy is much more sample-efficient, as observed by the higher deltas between successive experiments, especially in the range 250-500 and 500-750 demos.
Notably, our policy achieves an average success rate across tasks of $\sim 60 \%$ with 500 demos, while the image policy is slightly above $30 \%$.
This result clearly showcases the sample-efficiency benefits of keypoint representations over RGB images.

\begin{table}[!t]
     \vspace{-2.5mm}
    \centering
    \resizebox{\textwidth}{!}{%
 %   \begin{adjustbox}{width=.5\textwidth,center}

    \begin{tabular}{lccccc}
    \toprule
    % \multirow{2}{1em}{\textbf{Policy}} & \multirow{2}{2em}{\textbf{Gen.}} & \multirow{2}{2em}{{Threading}} & \multirow{2}{2em}{{Piece}}  & \multirow{2}{2em}{{Box}} & \multirow{2}{1em}{{Tray}} \\
    %  & \textbf{Data.} & & {Assembly} & {Cleanup} & {Lift } \\
     \textbf{Policy} & \textbf{Gen.Data} & Threading & {PieceAssembly} & {BoxCleanup} & {TrayLift} \\
    \midrule
     RGB \textit{(scratch)}  & DexMimicGen \cite{DexMimicGen} & $45.0 \pm \small 3.7$ & $76.0 \pm \small 4.5$ & $86.3 \pm \small 1.7$ & $62.3 \pm \small 2.0$  \\
     RGB-D \textit{(scratch)} & DexMimicGen \cite{DexMimicGen}  & $35.3 \pm \small 3.3$ & $77.3 \pm \small 2.8$ & $86.6 \pm \small 1.8$ & $65.0 \pm \small 2.1$  \\
     RGB \textit{(SD-DINOv2)}  & DexMimicGen \cite{DexMimicGen} & $45.3 \pm \small 3.3$ & $77.0 \pm \small 3.2$ & $86.3 \pm \small 1.7$ & $62.0 \pm \small 2.1$  \\
    %  MAC & 0 & 0 & 0 & 0 & 0 & 0 \\
     \midrule 
     Keypoint \textit{(SD-DINOv2)} & Ours & $\mathbf{73.9 \pm \small 2.1}$ & $\mathbf{84.6 \pm \small 0.5}$ & $\mathbf{94.6 \pm \small 0.6}$ & $\mathbf{91.6 \pm \small 0.8}$  \\
     \bottomrule
    \end{tabular}%
  }
 %  \end{adjustbox}
    \caption{Image vs. keypoint policy success rates (\%) in 4 simulation tasks over 3 source demo and test seeds.}
    \label{tab:sim_1}
    \vspace{-1mm}
\end{table}

\noindent \textbf{Comparison with image policies} We conduct experiments for all aforementioned baselines for all four tasks, using $1000$ fixed initial test scenes for each of the three source demo seeds as before.
We then compare the final success rates of policies trained in DexMimicGen's and our's datasets, for the total amount of demos that each method is able to generate within a fixed collection time budget.
In particular, using the calculation explained previously, our policy generates $>10^6$ augmentations within the same collection budget as DexMimicGen (33 min.).
We however use early stopping when training our policy to avoid diminishing returns due to overfitting to the augmentations \cite{DexMimicGen}, which we observe is at average between 5 and 10 thousand demos.
Results are reported in Tab.~\ref{tab:sim_1}.
Our policy far exceeds the performance of all image-based baselines, with highest margin in the \textit{Threading} task ($28.9\%$ delta).

\begin{figure*}[!t]
  \centering
  \includegraphics[width=\textwidth]{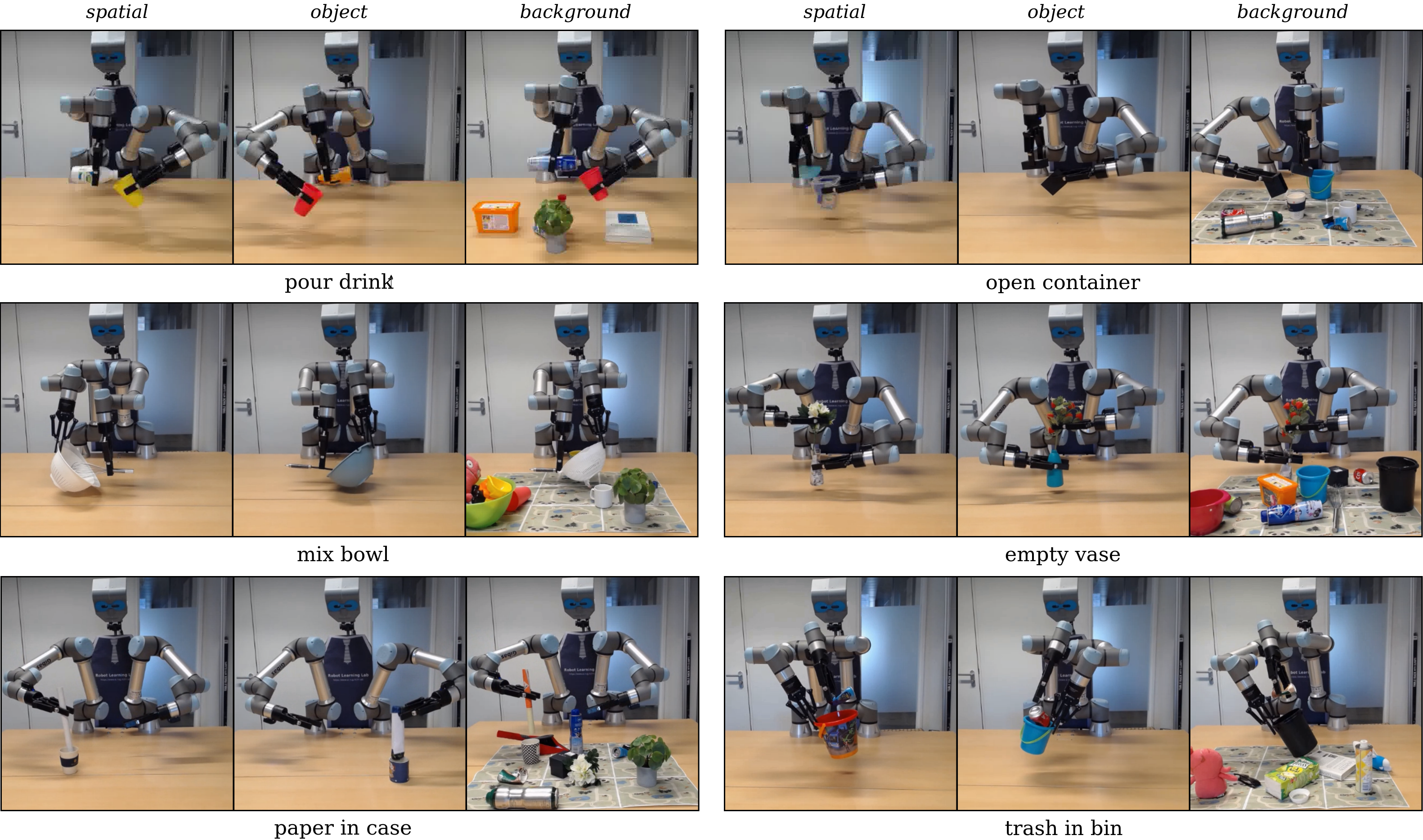}
  \caption{\textbf{Test tasks with a real robot.} We conduct extensive experiments in 6 diverse tasks, containing different object categories, manipulation skills and arm-coordination strategies. For each task, we explicitly test for generalization in novel spatial arrangements, object instances and background scene noise.}
  \label{fig:real-viz}
    % \vspace{-2mm}
\end{figure*}

\subsection{Real Robot Experiments}
\label{exp:real}

\begin{figure}[!t]
  \centering
  \includegraphics[width=\textwidth]{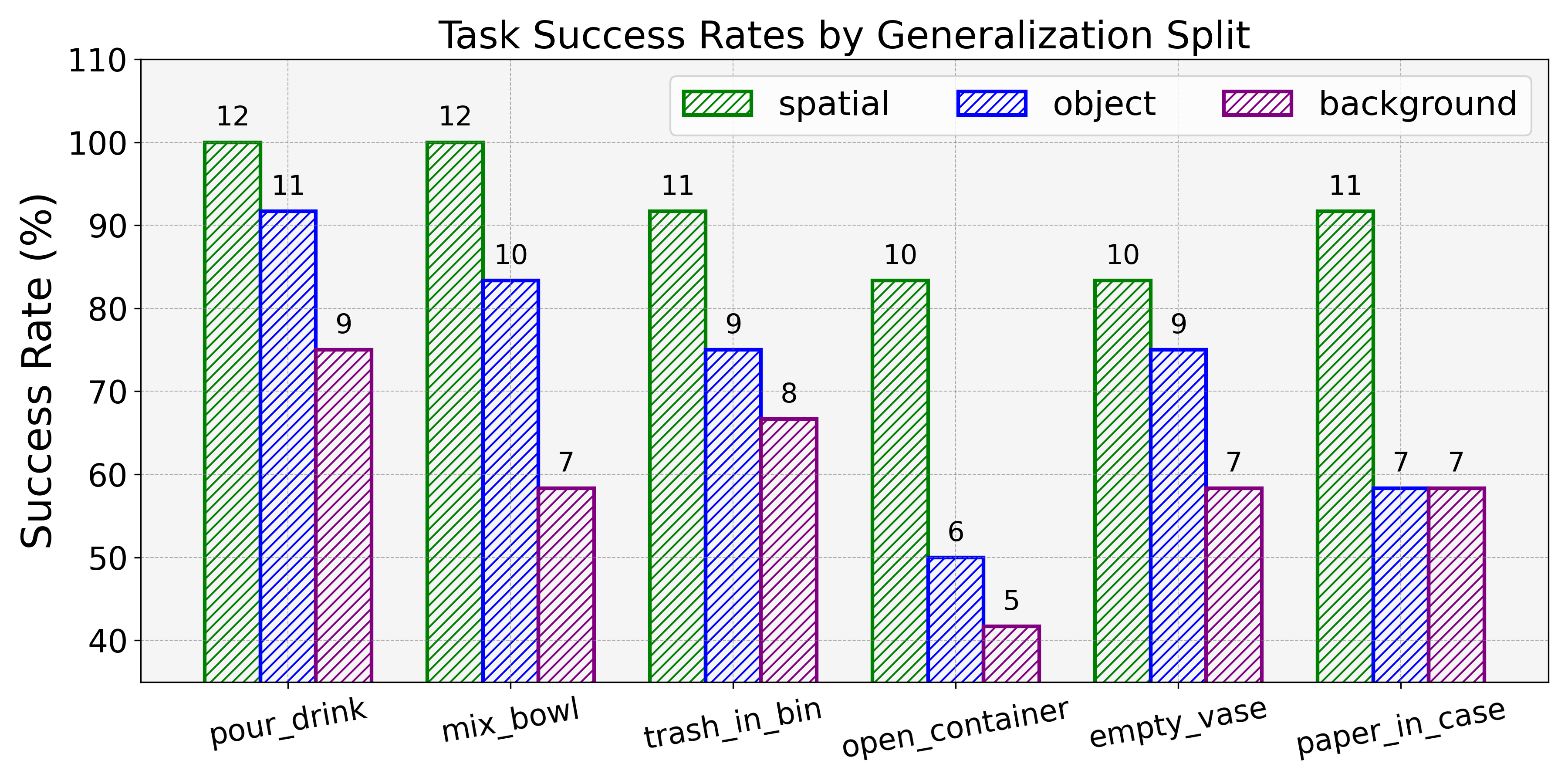}
  \caption{\textbf{Generalization experiments in 6 real-world tasks.} We evaluate for unseen \textit{spatial} arrangements, \textit{object} instances and \textit{background} noise. We conduct 12 trials per task-scenario combination and report the total number of successes on top of each bar.}
  \label{fig:real-exp}
    \vspace{-2mm}
\end{figure}

We conduct experiments on 6 bimanual real-world tasks: (a) \textit{pour drink}, picking a bottle and a cup with each arm and pouring from the bottle to the cup, (b) \textit{mix bowl}, picking a whisk and a bowl with each arm and mixing the bowl with the whisk, (c) \textit{trash in bin}, picking a trash soda can and a bin with each arm and dropping the trash in the bin, (d) \textit{empty vase}, picking a vase with flowers with one arm and removing the flowers with the other, (e) \textit{open container}, picking a container with one arm and removing it's lid with the other, and (f) \textit{paper in case}, picking a paper roll with one arm, handing it over to the other arm and then placing it inside a pencil case.
Our tasks are chosen to represent a broad variety of manipulation skills, different arm coordination strategies and diverse object categories.
For each task, we explicitly collect object instances of the same category with variations in appearance and shape to perform generalization experiments.
An illustration of task trials is shown in Fig.~\ref{fig:real-viz}.

\noindent \textbf{Generalization} We evaluate for all 6 tasks in three generalization splits: (a) \textit{spatial}, containing unseen arrangements of the demo objects, (b) \textit{object}, containing unseen object instances in unseen arrangements, and (c) \textit{background}, containing cluttered scenes with random background objects and both seen/unseen object instances in unseen arrangements.
We perform 12 trials per task and split, for a total of 216 trials.
Results are shown in Fig.~\ref{fig:real-exp}.
\begin{table}[!t]
     \vspace{-2.5mm}
    \centering
    \resizebox{\textwidth}{!}{%
 %   \begin{adjustbox}{width=.5\textwidth,center}

    \begin{tabular}{lcccc}
    \toprule
    \textbf{Method} & \textbf{\#Params} & {pour\_drink} & {mix\_bowl}  & {trash\_in\_bin} \\
    \midrule
    R+X (\textit{gpt-4o}) \cite{R+X} & - & 6/9 & 3/9 & 5/9  \\
    KALM-diffuser  \cite{KALM} & $56.6M$ & {7/9} & \textbf{8/9} & 6/9  \\
    Kp-RDT  (Ours) & $5.7M$ & \textbf{8/9} & \textbf{8/9} & \textbf{7/9}  \\
    %  MAC & 0 & 0 & 0 & 0 & 0 & 0 \\
     \bottomrule
    \end{tabular}%
  }
 %  \end{adjustbox}
    \caption{Trajectory-level policy success rates in 3 real-world tasks. }
    \label{tab:real_comparisons}
    \vspace{-1mm}
\end{table}

\begin{figure}[!t]
  \centering
  \includegraphics[width=\textwidth]{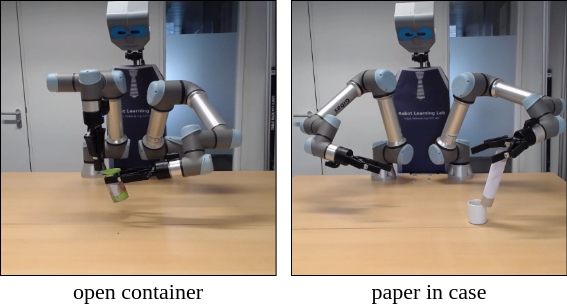}
  \caption{\textbf{Most common failure cases:} In high-precision tasks such as removing a thin container lid or placing a thick paper roll in a small pencil case, the policy tends to fail in unseen object instances when they significantly vary in shape from the demo objects.}
  \label{fig:real-fail}
    % \vspace{-2mm}
\end{figure}

We observe that our trained policies are robust in spatial arrangements, with an average success rate of $91.6 \%$, showcasing the effectiveness of our spatial augmentations.
In novel object instances, the average success rate is $72.2 \%$, with the policy mostly failing in the \textit{open container} and \textit{paper in case} tasks, which is due to considerable change in shape between the demo-test objects.
This leads to failures in high-precision tasks such as removing a small container lid or placing the paper in a small case.
We visually verify that keypoint correspondences are in all cases (reasonably) sound, so this failure mode comes from the policy's inability to handle diverse shapes, due to the single demonstration used.
In the other 4 tasks, similar issues are much less prominent, as the demonstrated behavior suffices to cover the shape distribution of different objects. 
Overall, we find our results promising with regards to the use of keypoints as underlying state representations.
In the background split, the average success rate is $59.7 \%$, with most failures due to infeasible grasps and collisions with neighbouring objects. 
In some cases we also observed floaters from the point tracker, as an artifact of loss of visibility due to scene clutter and access to only single-view. 
We note however that the reported performance gap in this split is expected, considering that the policy has not been distilled from obstacle avoidance-integrated TAMP and relies simply on IK motion planning.
Some examples of failure cases are illustrated in Fig.~\ref{fig:real-fail}.

\noindent \textbf{Comparisons with trajectory-level policies} We also investigate the performance of our Kp-RDT policy, when compared to recent works that rely on predicting action trajectories from keypoint-based representations.
We use two recent works: (a) \textbf{R+X} \cite{R+X}, which uses an LLM with keypoint-action token prompting strategy \cite{Palo2024KeypointAT} and in-context examples, and (b) \textbf{KALM-diffuser} \cite{KALM}, which trains a keypoint-conditioned variant of Diffuser-Actor \cite{DiffuserActor} for trajectory prediction.
Both these works predict the entire action trajectory from the initial keypoint state and execute it open-loop.
We use \textit{gpt-4o} \cite{GPT4} for R+X, the UNet implementation from the authors for KALM \cite{KALM} and use $5000$ demo augmentations for all methods.
We follow the same setup with Kp-RDT as in KALM and train trajectory-level policies with $H=96$ and $100$ diffusion steps. 
For R+X, we compute MSE between the observed keypoint coordinates and those in memory (from the augmentations) and retrieve the top-3 most similar for in-context prompting.

We conduct experiments in 3 of our tasks (\textit{pour drink, mix bowl} and \textit{trash in bin}), with 3 trials per task-split, for a total of 27 trials per method, where we set the initial object arrangement as close as possible.
Results are given in Tab.~\ref{tab:real_comparisons}.
The zero-shot LLM policy achieves an average success rate of $51.8 \%$, with most failures in the mixing task due to inability to grasp the whisk under certain orientations.
These results showcase the necessity for distilling the augmentations with a trained policy, instead of relying on LLM in-context interpolation.
Both KALM-diffuser and our method achieve similar results $\sim 80 \%$, although our method is an order of magnitude smaller in raw parameter count.
This is because unlike KALM, our Kp-RDT architecture does not require SD-DINO feature maps and projectors for them for conditioning the policy, instead only relying on 3D keypoint coordinates and group embeddings to encode semantics.

\section{CONCLUSIONS}
In this work we present \textit{PAD}, a unified framework for learning bimanual visuomotor policies from a single human video demonstration.
Our framework proposes to use keypoints as underlying state representations to parse the video into robot-executable data, augment the data at scale in a simulation-free fashion and distill the augmented data with a diffusion policy.
We design a bimanual TAMP procedure for specializing demo augmentations to the bimanual case and propose Kp-RDT, an adapted version of RDT \cite{RDT} that supports keypoint conditioning.
Empirically, we show that our framework is superior to state-of-the-art bimanual demo augmentation methods relying on simulation rollouts, in terms of success rates, data collection time and sample-efficiency.
We apply our framework in six real-world tasks and show that PAD obtains policies that generalize to unseen spatial arrangements, object instances and background scene noise, while doing so from a single human demonstration.

\noindent \textbf{Limitations \& Future Work} As discussed earlier, our policies have a considerable error rate at high-precision tasks for unseen object instances that significantly vary in shape from the demo object.
This can be alleviated by using more training videos, for demonstrating how to adapt the policy in cases of varying shapes.
Second, our TAMP procedure and final policies currently do not consider obstacle avoidance, which can be integrated in the future via using third-party motion planners and some keypoint representation for obstacles.
Third, our augmentation framework currently relies on the 3D rigidity assumption, and hence doesn't support deformable objects.
Finally, in its current form our work only supports training single-task policies from separate videos.
A future avenue will explore combining keypoint with language-conditioning to obtain multi-task policies that can follow text instructions, while still enjoying the generalization benefits of keypoint abstractions.


%%%%%%%%%%%%%%%%%%%%%%%%%%%%%%%%%%%%%%%%%%%%%%%%%%%%%%%%%%%%%%%%%%%%%%%%%%%%%%%%

%%%%%%%%%%%%%%%%%%%%%%%%%%%%%%%%%%%%%%%%%%%%%%%%%%%%%%%%%%%%%%%%%%%%%%%%%%%%%%%%



% \section*{ACKNOWLEDGMENT}

% The preferred spelling of the word ÒacknowledgmentÓ in America is without an ÒeÓ after the ÒgÓ. Avoid the stilted expression, ÒOne of us (R. B. G.) thanks . . .Ó  Instead, try ÒR. B. G. thanksÓ. Put sponsor acknowledgments in the unnumbered footnote on the first page.

%%%%%%%%%%%%%%%%%%%%%%%%%%%%%%%%%%%%%%%%%%%%%%%%%%%%%%%%%%%%%%%%%%%%%%%%%%%%%%%%

\bibliographystyle{unsrt}
\bibliography{main}

\end{document}